\documentclass[conference]{IEEEtran}
\IEEEoverridecommandlockouts

\usepackage{cite}
\usepackage{amsmath,amssymb,amsfonts}
\usepackage{graphicx}
\usepackage{textcomp}
\usepackage{xcolor}
\usepackage{float}
\usepackage{algorithm}

\usepackage{algpseudocode}
\usepackage{caption}
\usepackage[numbers,sort&compress]{natbib}
\usepackage[caption=false,font=footnotesize]{subfig}
\usepackage{tcolorbox}
\tcbuselibrary{listingsutf8}
\usepackage{listings}
\usepackage{multirow}
\usepackage{booktabs}
\usepackage{multirow}

\def\BibTeX{{\rm B\kern-.05em{\sc i\kern-.025em b}\kern-.08em
    T\kern-.1667em\lower.7ex\hbox{E}\kern-.125emX}}

\begin{document}

\title{PCBnet: A Dataset and Automatic Constructing of
SPICE Netlists from Schematic Images}
\author{
\IEEEauthorblockN{
Zhen Huang\textsuperscript{1,3,4,*},
Yuhao Gao\textsuperscript{2,*},
Yuzhi Liu\textsuperscript{2},
Daian Cheng\textsuperscript{2},
Chengyuan Shao\textsuperscript{2},
Yucheng Chen\textsuperscript{2},
Yongjian Jia\textsuperscript{2}, \\ 
Futing Zhang\textsuperscript{2},
Yichen Shi\textsuperscript{1},
Wenhao Wang\textsuperscript{2},
Zuyan He\textsuperscript{2}, 
Yangbo Wei\textsuperscript{1},
Zhanfei Chen\textsuperscript{2}, \\
Jinlong Yan\textsuperscript{2},
Yu Zhang\textsuperscript{3},
Haoying Wu\textsuperscript{5},
Ting-Jung Lin\textsuperscript{1,2,4,$\dagger$},~Lei He\textsuperscript{1,2,4,$\dagger$}
}

\IEEEauthorblockA{
\textsuperscript{1}Eastern Institute of Technology, Ningbo, China\\
\textsuperscript{2}Ningbo Institute of Digital Twin, Eastern Institute of Technology, Ningbo, China\\
\textsuperscript{3}University of Science and Technology of China, Hefei, China\\
\textsuperscript{4}Engineering Research Center of Chiplet Design and Manufacturing of Zhejiang Province\\
\textsuperscript{5}Wuhan University of Technology, Wuhan, China
}

\thanks{\textsuperscript{*}Equal contribution.}
\thanks{\textsuperscript{$\dagger$}Corresponding authors: Ting-Jung Lin (tlin@idt.eitech.edu.cn), Lei He (lhe@eitech.edu.cn).}
}

\maketitle

\begin{abstract}
PCBs are fundamental to modern electronic systems, yet AI-driven PCB design automation remains constrained by the lack of large-scale paired schematic--netlist datasets. PCB schematics are particularly challenging due to diverse component types, complex wiring topologies, and noisy textual annotations.

To address this gap, we present PCBnet, the first large-scale PCB schematic dataset, comprising over 300 real-world designs with annotated pins and SPICE netlists. It contains more than \textbf{50,000} component instances, \textbf{150,000} wires, \textbf{100,000} text regions, and \textbf{400,000} characters. We further develop an automated schematic-to-netlist pipeline that combines visual recognition with domain-knowledge-guided multi-agent correction. The proposed method achieves 94.54\% component detection mAP, 98.57\% text recognition accuracy, and 84.47\% end-to-end connectivity accuracy. PCBnet provides a benchmark and data foundation for future AI-driven PCB design automation.

\end{abstract}


\section{Introduction}

PCBs are the hardware foundation of modern intelligent systems, including autonomous vehicles, robotics, AI accelerators, and edge devices. However, PCB design remains largely manual due to diverse component types, complex wiring topologies, and heterogeneous design conventions, creating a major bottleneck in electronic product development.

AI has shown strong potential in Electronic Design Automation (EDA)~\cite{chaudhuri2025latent, wei2026vflow, huang2025selfattention, wei2025modelgen}. For integrated circuits (ICs), recent works such as ChipNemo~\cite{ref18} demonstrate that foundation models trained on high-quality domain-specific data can capture design knowledge and support automated workflows. In analog and mixed-signal (AMS) design, the AMSnet series~\cite{ref14,ref15,ref16,ref17} further shows the value of multimodal datasets and structured knowledge for circuit understanding and generation.

\begin{figure}[!htbp]
\centering
\includegraphics[width=\linewidth]{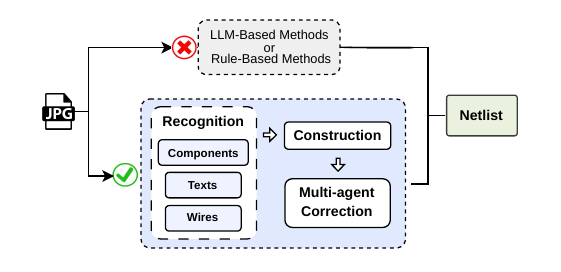}
\caption{Overview of PCBnet: the proposed pipeline reconstructs circuit netlists from PCB schematics through visual recognition, topology construction, and multi-agent correction.}
\label{fig:pipeline}
\vspace{-5mm}
\end{figure}

Despite these advances, AI-driven PCB design still lacks large-scale structured data. Existing PCB data is scattered across datasheets, evaluation-board documents, and open-source repositories, often in heterogeneous and unstructured formats. Public PCB datasets are also limited, and most of them, such as DeepPCB, HRIpcb~\cite{huang2020hripcb}, and related datasets~\cite{huang2019pcb, lv2024dataset}, focus on defect detection rather than schematic understanding. As a result, there is still no large-scale schematic--netlist paired dataset for modeling PCB circuit topology.

To address this gap, we present \textbf{PCBnet}, a large-scale dataset for PCB schematic understanding and netlist construction. PCBnet contains over 300 real-world PCB designs with multi-level annotations, including components, pins, wires, texts, and SPICE netlists. It covers more than \textbf{50,000} component instances, \textbf{150,000} wire annotations, \textbf{100,000} text regions, and \textbf{400,000} labeled characters.

As shown in Fig.~\ref{fig:pipeline}, we further develop a schematic-to-netlist pipeline that integrates visual element recognition, topology construction, and multi-agent correction with domain knowledge. On PCBnet, our method achieves \textbf{94.54\%} component detection mAP, \textbf{98.57\%} text recognition accuracy, and \textbf{84.47\%} connectivity accuracy, demonstrating the effectiveness of PCBnet as both a dataset and benchmark for AI-driven PCB design automation.

The main contributions of this work are summarized as follows.
\begin{itemize}
\item \textbf{PCBnet Dataset.} We introduce the first large-scale PCB schematic dataset with detailed annotations of components, pins, wires, texts, and paired SPICE netlists.

\item \textbf{Schematic-to-Netlist Framework.} We propose a pipeline that combines visual recognition, structure construction, and multi-agent correction to recover circuit connectivity from schematic images.

\item \textbf{PCB Schematic Benchmark.} We establish a benchmark for component detection, text recognition, and netlist construction, providing a data foundation for future AI-driven PCB design research.
\end{itemize}

\begin{figure*}[t]
    \centering
    \includegraphics[width=\textwidth]{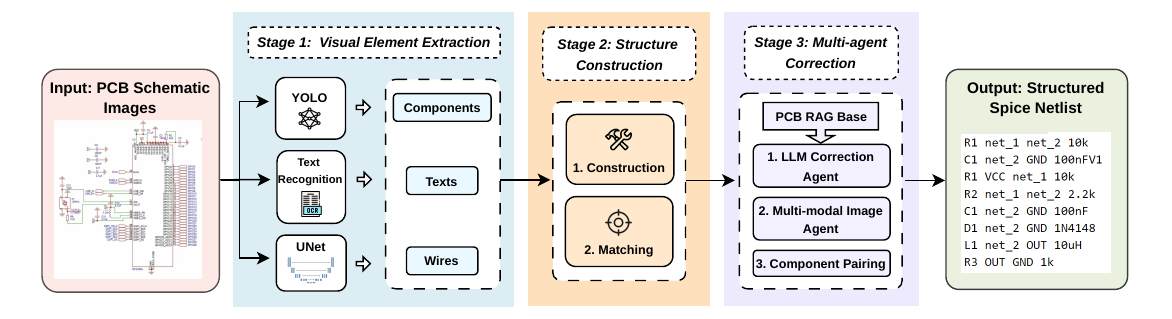}
    \caption{Overview of the schematic-to-netlist pipeline. Stage 1 extracts visual elements, including components, wires, and texts, from input PCB schematics. Stage 2 performs structure construction and element matching. Stage 3 applies a multi-agent correction framework, supported by a knowledge base via RAG, to correct and complete the final structured SPICE netlist. }
    \label{fig:framework}
    \vspace*{-5mm}
\end{figure*}

\section{Dataset Construction}

\subsection{Dataset Collection and Annotation}

PCBnet is constructed from over 300 real-world PCB schematic files collected from open-source hardware projects and EDA tools, including KiCad and EasyEDA. The designs cover diverse circuit types, component categories, wiring structures, and schematic styles.

For each design, we render the schematic image and extract the corresponding SPICE netlist from the original engineering file, ensuring one-to-one image--netlist correspondence. Each schematic is annotated with component bounding boxes, pin locations, text regions, and wire masks. Component and text annotations are initialized using pretrained models and manually corrected for accuracy.

\subsection{Connectivity Representation and Statistics}

Each schematic is represented as a circuit graph
\begin{equation}
G=(V,E),
\end{equation}
where $V$ denotes component pins and $E$ denotes electrical connections. Let $p_{i,k}$ be the $k$-th pin of component $i$ and $w_m$ a wire segment. A pin--wire connection is established if
\begin{equation}
d(p_{i,k},w_m)<\epsilon,
\end{equation}
where $\epsilon$ is a distance threshold. The resulting pin--wire associations are used to construct the circuit graph and generate the SPICE netlist.

\section{Image-to-Netlist Pipeline}

Given a PCB schematic image $I$, the framework reconstructs its circuit topology and generates a structured SPICE netlist. As shown in Fig.~\ref{fig:framework}, it consists of three stages: \textbf{Visual Element Extraction}, \textbf{Circuit Structure Construction}, and \textbf{Multi-agent Correction}.

\subsection{Stage 1: Visual Element Extraction}

\textbf{Component Detection.}
YOLOv11 detects the component set $C={c_i}$, where each component is represented by a bounding box $b_i$ and category label $l_i$.

\textbf{Text Recognition.}
Text regions are localized by YOLOv11 and recognized by PaddleOCR~\cite{PaddleOCR}. For vertically oriented labels, OCR is applied to both the original crop and its $90^\circ$-rotated version, and the higher-confidence result is selected. The recognized text set is denoted as
\[
T = \{(t_i, b_i, c_i)\},
\]
where $t_i$ is the recognized token, $b_i$ is the text bounding box, and $c_i$ is the OCR confidence score.

\subsubsection{Wire Segmentation}

Wire structures are extracted using the original U-Net architecture with default settings. 
Given the input image $I$, the network predicts a binary wire mask $M = Seg(I)$, where $M(x,y)=1$ indicates a wire pixel. 
We further apply skeletonization to obtain a one-pixel-wide representation of the wiring topology, denoted as $W=\{w_k\}$.

\subsection{Stage 2: Structure Construction}\label{stage:2}

After obtaining components $C$, texts $T$, and wires $W$, Stage 2 constructs the circuit topology through two steps: \textit{geometric construction} and \textit{text-component matching}.

\subsubsection{Construction}

We first construct the geometric structure of the circuit by estimating pin locations and inferring electrical connectivity.

\textbf{Pin Estimation.}
Pin locations are estimated based on component categories and geometric layouts. Common component types, such as resistors, capacitors, and integrated circuits, typically exhibit consistent pin configurations, allowing candidate pin locations to be derived in a geometry-guided manner.

\textbf{Connectivity Inference.}
We infer electrical connections by analyzing the spatial relationships between estimated component pins and wire segments. Each pin is treated as a node, and a connection is established when a pin is spatially adjacent to a wire segment. The resulting circuit topology is represented as a graph $G=(V,E)$, where $V$ denotes component pins and $E$ denotes the inferred electrical connections.

\subsubsection{Text-Component Matching}

After constructing the geometric topology, we associate textual labels with their corresponding components. The matching process combines spatial proximity with schematic-specific naming conventions and layout characteristics.

For each text instance $t_i$, we first identify candidate components in its local neighborhood and obtain an initial association using nearest-neighbor matching based on Euclidean distance. This approach is effective for regular schematic layouts, where component identifiers are generally placed close to their corresponding components.

However, dense layouts, wire occlusions, and flexible text placement can make nearest-neighbor matching ambiguous. We therefore refine the initial associations using three lightweight constraints:

\begin{itemize}
\item \textbf{Category Consistency Constraint:} Text prefixes such as `R'', `C'', and ``U'' are required to agree with the corresponding component categories, reducing cross-type mismatches.
\item \textbf{Geometric Alignment Constraint:} Text-component pairs with horizontal or vertical alignment are prioritized according to common schematic layout conventions.
\item \textbf{Local Uniqueness Constraint:} Conflicting assignments are reduced by preventing multiple text instances within a local region from being matched to the same component.
\end{itemize}

By combining distance-based matching with these structured constraints, the proposed method improves the robustness of text-component association. The matched components, pins, wires, and textual labels are then organized into a structured circuit topology and converted into a SPICE netlist.

\subsection{Stage 3: Multi-agent Correction}

Stage 2 produces an initial circuit topology, but recognition errors, particularly in text, may propagate into structural inaccuracies such as misidentified components or incorrect net labels.

OCR errors often follow common patterns, including missing underscores and confusion between visually similar characters (e.g., \texttt{1/l/I} and \texttt{O/0}). Since these errors are difficult to resolve through visual processing alone, Stage 3 introduces a multi-agent correction framework that integrates recognition results with domain-specific knowledge.

\subsubsection{Confidence-Guided Gatekeeper}
A gatekeeper module first filters OCR outputs based on confidence. Predictions with $c_i \ge \tau$ are accepted directly, while low-confidence tokens are routed to the correction module. This design focuses computation on uncertain cases and avoids unnecessary overhead.

\subsubsection{LLM-based Correction with Domain Knowledge}
For low-confidence tokens, we employ an LLM-based correction agent augmented with a lightweight domain knowledge base. The knowledge base encodes naming conventions, common character confusion patterns, and representative correction examples observed in PCB schematics.

Given an OCR token $t_i$, the system retrieves relevant rules $\mathcal{R}_i$ and performs constrained generation:
\[
\hat{t}_i^{(L)} = \arg\max_{t} \; P(t \mid t_i, \mathcal{R}_i),
\]
where $\hat{t}_i^{(L)}$ denotes the corrected token. This design improves correction accuracy while restricting outputs to valid naming patterns.

\subsubsection{Multi-modal Image correction}
When rule-based correction is insufficient, we introduce a multimodal correction step that directly leverages visual information. The cropped text image $I_i$ is fed into a vision-language model (e.g., GPT-4.1 / GPT-5.4) to produce a candidate:
\[
\hat{t}_i^{(V)} = f_{\text{vlm}}(I_i).
\]

The final prediction is selected based on confidence:
\[
\hat{t}_i =
\begin{cases}
\hat{t}_i^{(V)}, & \text{if } c(\hat{t}_i^{(V)}) > c(t_i) \\
t_i, & \text{otherwise}.
\end{cases}
\]

This module is particularly effective for short text tokens, where visual models tend to be more reliable. Moreover, since most tokens are already correctly recognized in earlier stages, this correction is only applied to a small subset of cases and thus introduces limited additional overhead.

\subsubsection{Component-Aware Pairing}

To further reduce ambiguity, each low-confidence token is paired with its associated component obtained from the text-component matching in stage \ref{stage:2}. Specifically, we use the category of the matched component as a semantic constraint:
\[
\hat{t}_i^{(C)} = \arg\max_{t} \; P(t \mid t_i, l_j),
\]
where $l_j$ denotes the category of the corresponding component. This constraint (e.g., \texttt{R} for resistors, \texttt{C} for capacitors) improves correction accuracy in ambiguous cases.

Finally, by integrating visual element extraction, Structure construction, and multi-agent correction, the proposed framework enables reliable and consistent schematic-to-netlist construction.

\section{Experiments}
\subsection{Experimental Setup and Evaluation}

We split PCBnet into training, validation, and test sets at the design level with a ratio of 8:1:1. All experiments are implemented in PyTorch and conducted on a workstation with an NVIDIA RTX 3090 GPU and an Intel Core i9 processor.

We evaluate the full pipeline on three tasks:
\begin{itemize}
\item \textbf{Component Detection}, evaluated by mean Average Precision (mAP);
\item \textbf{Text Recognition}, evaluated by Character Error Rate (CER), Word Error Rate (WER), and accuracy (ACC);
\item \textbf{Netlist Construction}, evaluated by connectivity accuracy, i.e., the proportion of correctly recovered pin-to-pin connections.
\end{itemize}

\subsection{Component Detection Results}

We first evaluate component detection performance on PCBnet using mAP. Our method achieves \textbf{94.54\% mAP}, demonstrating strong detection performance across diverse PCB schematic layouts.

PCB schematics contain substantial variation in component types, symbol styles, scales, and orientations. In addition, real-world layouts are often dense and contain overlapping visual elements, making accurate localization nontrivial. The high mAP shows that the proposed detection module can reliably localize and classify components under these challenging conditions.

In contrast, even when provided with \textbf{cropped component regions}, general-purpose multimodal models (e.g., GPT-5.4, Claude Opus 4.6, and Gemini 3 Pro) achieve substantially lower accuracy (around 30\%). When applied directly to full schematic images, these models often fail to reliably localize individual components due to the dense and fine-grained layout, highlighting their limitations in precise localization and structured visual parsing required for schematic understanding.

\begin{table}[t]
\setlength{\tabcolsep}{4pt}
\centering
\caption{Comparison of component detection, wire extraction, and netlist construction accuracy.}
\label{tab:netlist_comparison}
\begin{tabular}{lccc}
\hline
Method & Comp. mAP & Wire. IoU & Conn. Acc. \\
\hline
\textbf{Ours (PCBnet)} & \textbf{94.54} & \textbf{93.70} & \textbf{84.47} \\
\hline
GPT-5.4 & 34.28 & -- & 8.74 \\
Claude Opus 4.6 & 29.63 & -- & 7.91 \\
Gemini 3 Pro  & 33.05 & -- & 9.26 \\
DeepSeek-V3.2 & 27.17 & -- & 7.02 \\
Qwen3.5-Plus & 29.44 & -- & 8.01 \\
\hline
\end{tabular}
\end{table}

\subsection{Text Recognition Results}

Table~\ref{tab:ocr_accuracy} compares different OCR engines on the PCB schematic dataset, including Tesseract~\cite{Tesseract}, EasyOCR~\cite{easyocr2020}, MMOCR~\cite{MMOCR}, and DocTR~\cite{DocTR}. We consider two settings: \textit{Full Image Mode}, where OCR is applied directly to the entire schematic, and \textit{YOLO+ Mode}, where text regions are first localized before recognition.

\begin{table}[!t]
\caption{Comparison of OCR engines under full-image and detection-assisted settings.}
\begin{center}
\begin{tabular}{llccc}
\toprule
\textbf{Mode} & \textbf{Method} & \textbf{CER (\%)} & \textbf{WER (\%)} & \textbf{ACC (\%)} \\
\midrule
\multirow{5}{*}{Full Image}
& PaddleOCR & 21.35 & 37.40 & 78.65 \\
& EasyOCR & 60.34 & 94.68 & 39.66 \\
& Tesseract & 22.34 & 42.16 & 77.66 \\
& DocTR & 22.99 & 50.00 & 77.01 \\
& MMOCR & 39.02 & 50.77 & 60.98 \\
\midrule
\multirow{6}{*}{YOLO+}
& PaddleOCR & \textbf{6.33} & \textbf{19.04} & \textbf{93.67} \\
& EasyOCR & 49.90 & 85.83 & 50.10 \\
& Tesseract & 50.84 & 70.80 & 49.16 \\
& DocTR & 13.86 & 32.40 & 86.14 \\
& MMOCR & 16.61 & 33.10 & 83.39 \\
& TrOCR & 13.26 & 35.95 & 86.74 \\
\bottomrule
\end{tabular}
\end{center}
\label{tab:ocr_accuracy}
\end{table}

\begin{table}[!t]
\caption{Ablation study of the multi-agent correction modules.}
\begin{center}
\begin{tabular}{lccc}
\toprule
\textbf{Method} & \textbf{CER (\%)} & \textbf{WER (\%)} & \textbf{ACC (\%)} \\
\midrule
YOLO+PaddleOCR & 6.33 & 19.04 & 93.67 \\
+ Knowledge-guided LLM & 4.07 & 8.45 & 95.93 \\
+ Multimodal Image Agent & 2.52 & 4.67 & 97.48 \\
+ Component Pairing & 1.43 & 3.59 & 98.57 \\
\bottomrule
\end{tabular}
\end{center}
\label{tab:agent_accuracy}
\end{table}

All methods perform poorly in the Full Image setting, with CER typically above 20\% and WER above 40\%. This indicates that directly applying OCR to full schematic images is highly susceptible to background noise, dense layouts, and non-text elements.

By contrast, the YOLO+ setting significantly improves performance across all methods, showing that accurate text localization is important for robust schematic OCR. Among them, YOLO+PaddleOCR achieves the best performance, with \textbf{6.33\% CER}, \textbf{19.04\% WER}, and \textbf{93.67\% ACC}. This result confirms the effectiveness of decoupling text detection and recognition in complex schematic images.

To further improve recognition performance, we incrementally add each module of the proposed multi-agent correction mechanism to the YOLO+PaddleOCR baseline. The results are shown in Table~\ref{tab:agent_accuracy}.

Each module consistently improves recognition performance. The knowledge-guided LLM reduces CER from 6.33\% to 4.07\%, showing that domain-specific rules can effectively resolve common OCR ambiguities. The multimodal image agent further lowers CER to 2.52\%, indicating that visual information is beneficial for handling ambiguous or degraded text regions. Finally, component-aware pairing reduces CER to \textbf{1.43\%} and WER to \textbf{3.59\%}, achieving an overall accuracy of \textbf{98.57\%}.


\subsection{Netlist Construction Results}

Table~\ref{tab:netlist_comparison} compares the end-to-end schematic-to-netlist construction results. Our method achieves \textbf{93.70\%} wire extraction accuracy and \textbf{84.47\%} connectivity accuracy, where connectivity accuracy measures the proportion of correctly recovered pin-to-pin connections.

In comparison, general-purpose multimodal models achieve approximately \textbf{30\%} component accuracy and below \textbf{10\%} connectivity accuracy. These results highlight their difficulty in capturing the fine-grained geometric and topological structures of PCB schematics. By explicitly constructing circuit topology from recognized components, wires, and their structural relationships, our framework provides substantially more reliable netlist reconstruction.




\section{Conclusion}

In this work, we present \textbf{PCBnet}, a large-scale dataset for PCB schematic understanding and schematic-to-netlist construction. PCBnet provides multi-level annotations for components, pins, wires, texts, and paired SPICE netlists, covering over \textbf{50k} component instances, \textbf{150k} wires, \textbf{100k} text regions, and \textbf{400k} labeled characters.

We further develop a topology-oriented image-to-netlist framework that integrates visual recognition, structure construction, and multi-agent correction to recover circuit connectivity. Experiments show that our method achieves \textbf{94.54\%} component detection mAP, \textbf{98.57\%} text recognition accuracy, and \textbf{84.47\%} connectivity accuracy. PCBnet establishes a benchmark and data foundation for future AI-driven PCB design automation.



\clearpage

\raggedright
\bibliographystyle{IEEEtran}
\bibliography{reference}

\end{document}